\documentclass[journal,twocolumn]{IEEEtran}
\usepackage{slashbox,graphicx,times,amsmath,amssymb,cite,subfigure,stfloats,booktabs, url,multirow}
\usepackage[lined,ruled,commentsnumbered]{algorithm2e}
\allowdisplaybreaks

\makeatletter

\newcommand{\Rmnum}[1]{\expandafter\@slowromancap\romannumeral #1@}
\makeatother

\begin{document}
\title{On the Functional Equivalence of TSK Fuzzy Systems to Neural Networks, Mixture of Experts, CART, and Stacking Ensemble Regression}

\author{Dongrui~Wu,  Chin-Teng Lin, Jian Huang and Zhigang Zeng
\thanks{D.~Wu, J.~Huang and Z. Zeng are with the Key Laboratory of Image Processing and Intelligent Control (Huazhong University of Science and Technology), Ministry of Education, China. They are also with the School of Artificial Intelligence and Automation, Huazhong University of Science and Technology, Wuhan, China. Email: drwu@hust.edu.cn, huang\_jan@mail.hust.edu.cn, zgzeng@hust.edu.cn.}
\thanks{C-T Lin is with the Faculty of Engineering and Information Technology, University of Technology, Sydney, Australia. Email: Chin-Teng.Lin@uts.edu.au.}}

\maketitle

\begin{abstract}
Fuzzy systems have achieved great success in numerous applications. However, there are still many challenges in designing an optimal fuzzy system, e.g., how to efficiently optimize its parameters, how to balance the trade-off between cooperations and competitions among the rules, how to overcome the curse of dimensionality, how to increase its generalization ability, etc. Literature has shown that by making appropriate connections between fuzzy systems and other machine learning approaches, good practices from other domains may be used to improve the fuzzy systems, and vice versa. This paper gives an overview on the functional equivalence between Takagi-Sugeno-Kang fuzzy systems and four classic machine learning approaches -- neural networks, mixture of experts, classification and regression trees, and stacking ensemble regression -- for regression problems. We also point out some promising new research directions, inspired by the functional equivalence, that could lead to solutions to the aforementioned problems. To our knowledge, this is so far the most comprehensive overview on the connections between fuzzy systems and other popular machine learning approaches, and hopefully will stimulate more hybridization between different machine learning algorithms.
\end{abstract}


\IEEEpeerreviewmaketitle

\section{Introduction}

Rule-based fuzzy systems have achieved great success in numerous applications \cite{Mendel2017,Lin1996a}. There are two kinds of rules for a fuzzy system: \emph{Zadeh} \cite{Chang1972}, where the rule consequents are fuzzy sets, and \emph{Takagi-Sugeno-Kang (TSK)} \cite{Takagi1985}, where the rule consequents are functions of the inputs. Both types of fuzzy systems are universal approximators \cite{Buckley1993,Wang1992a}.

\begin{figure}[htbp]         \centering
\includegraphics[width=.9\linewidth,clip]{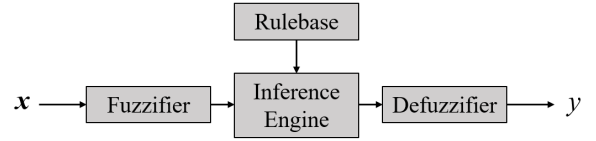}
\caption{Flowchart of a Zadeh (Mamdani) fuzzy system. The defuzzifier is not necessary in a TSK fuzzy system.} \label{fig:FLS}
\end{figure}

As shown in Fig.~\ref{fig:FLS}, a Zadeh (Mamdani) fuzzy system consists of four components: \emph{fuzzifier}, \emph{rulebase}, \emph{inference engine}, and \emph{defuzzifier}. The fuzzifier maps each crisp input into a fuzzy set, the inference engine performs inferences on these fuzzy sets to obtain another fuzzy set, utilizing the rulebase, and the defuzzifier converts the inferred fuzzy set into a crisp output. A TSK fuzzy system does not need the defuzzifier, because the output of the inference engine is already a crisp number. Because of their simplicity and flexibility, TSK fuzzy systems are much more popular in practice. This paper considers mainly TSK fuzzy systems for regression.

A TSK rule, as first proposed in \cite{Takagi1985}, assumes the following general form:
\begin{align}
\mbox{ If } f(x_1 \mbox{ is } X_1, ..., x_d \mbox{ is } X_d), \mbox{ then } y=g(x_1,...,x_d), \label{eq:Rule}
\end{align}
where $x_1,...,x_d$ are inputs, $X_1,...,X_d$ are membership functions (MFs), $y$ is the output, $f$ is a logical function connecting the antecedent propositions, and $g$ is a function of the inputs.

In practice, simple TSK rules are often preferred. As an example, a TSK fuzzy system with $d$ inputs and one output may have $K$ rules in the following form:
\begin{align*}
R_k: &\mbox{ If } x_1 \mbox{ is } X_{k,1} \mbox{ and } \cdots \mbox{ and } x_d \mbox{ is } X_{k,d}, \\
 &\mbox{ then } y_k(\mathbf{x}) = \sum_{i=1}^d a_{k,i}x_i + b_k, \quad k=1,...,K
\end{align*}
where $X_{k,i}$ is the MF for $x_i$ in the $k$th rule, and $a_{k,i}$ and $b_k$ are adjustable regression coefficients.

For a particular input $\mathbf{x}=(x_1,...,x_d)$, the membership grade on $X_{k,i}$ is $\mu_{X_{k,i}}(x_i)$, and the firing levels of the rule is:
\begin{align*}
f_k(\mathbf{x})=\mu_{X_{k,1}}(x_1)\times\cdots\times\mu_{X_{K,d}}(x_d).
\end{align*}


The output of the TSK fuzzy system is:
\begin{align}
y_{_{TSK}}(\mathbf{x})&=\frac{\sum_{k=1}^K f_k(\mathbf{x})\cdot y_k(\mathbf{x})}{\sum_{k=1}^K f_k(\mathbf{x})}\nonumber \\
&=\frac{\sum_{k=1}^K \left[f_k(\mathbf{x})\cdot\left(\sum_{i=1}^d a_{k,i}x_i+b_k\right)\right]}{\sum_{k=1}^K f_k(\mathbf{x})} \label{eq:yTSK1}
\end{align}
Or, if we define the normalized firing levels as:
\begin{align}
\bar{f}_k(\mathbf{x})=\frac{f_k(\mathbf{x})}{\sum_{k=1}^Kf_k(\mathbf{x})},\quad k=1,...,K \label{eq:f}
\end{align}
then, (\ref{eq:yTSK1}) can be rewritten as:
\begin{align}
y_{_{TSK}}(\mathbf{x})&=\sum_{k=1}^K\bar{f}_k(\mathbf{x})\cdot y_k(\mathbf{x}) \nonumber \\
&=\sum_{k=1}^K\left[\bar{f}_k(\mathbf{x})\cdot\left(\sum_{i=1}^d a_{k,i}x_i+b_k\right)\right] \label{eq:yTSK}
\end{align}


There are many challenges in designing an optimal fuzzy system, e.g.,
\begin{enumerate}
\item \emph{Optimization}. A fuzzy system can be optimized by evolutionary algorithms \cite{drwuEAAI2006}, gradient descent \cite{Wang1992b}, and gradient descent plus least squares estimation \cite{Jang1993}, as in the popular adaptive-network-based fuzzy inference system (ANFIS). However, evolutionary algorithms may be impractically slow on big data, gradient descent is very sensitive to the learning rate, and ANFIS can easily result in overfitting. So, it is necessary to develop more efficient and effective fuzzy system optimization algorithms, especially for big data.
\item \emph{Interpretability}. A well-known advantage of fuzzy systems over many other machine learning approaches is the interpretability, i.e., one can look at each rule and understand how the fuzzy system is working. However, the interpretability decreases when the number of rules increases, and when each input activates too many rules. How to increase the interpretability, without sacrificing the learning performance, is another challenge.
\item \emph{Curse of dimensionality}. Fuzzy systems are particularly suffering from the curse of dimensionality. Assume a fuzzy system has $d$ inputs, each with $p$ MFs in its domain. Then, the total number of rules is $p^d$, i.e., the number of rules increases exponentially with the number of inputs, and the fuzzy system quickly becomes unmanageable. Clustering could be used to reduce the number of rules (one rule is extracted for each cluster) \cite{Yager1994a,Delgado1997,Chiu1994,Juang1998}. However, the validity of the clusters also decreases with the increase of feature dimensionality, especially when different features have different importance in different clusters \cite{Jing2007,Jia2018}. Additionally, the high dimensionality of features also increases the number of antecedents in the rules, and hence make them difficult to interpret.
\item \emph{Generalization}. A fuzzy system should have not only good training performance, but also good generalization ability, i.e., it needs to perform well on the unknown test data. It is well-known in machine learning that regularization can improve the generalization performance; however, the concept of regularization has not been extensively explored in training fuzzy systems.
\end{enumerate}

This paper gives a comprehensive overview of the functional equivalence\footnote{It has been shown that many machine learning algorithms are universal approximators \cite{Wang1992a,Hornik1989}. However, two algorithms are both universal approximators does not mean that they are functionally equivalent: universal approximation usually requires a very large number of nodes or parameters, so it is theoretically important, but may not be used in real-world algorithm design. By functional equivalence, we emphasize that two algorithms can implement \emph{exactly} the same function with a relatively \emph{small} number of parameters.} of TSK fuzzy systems to four classical machine learning algorithms: neural networks \cite{Bishop1995}, mixture of experts (MoE) \cite{Jacobs1991}, classification and regression trees (CART) \cite{Breiman2017}, and stacking ensemble regression \cite{Zhou2012}. Although a few publications on the connections of TSK fuzzy systems to some of these approaches have scattered in the literature, to our knowledge, no one has put everything together in one place so that the reader can easily see the big picture and get inspired. Moreover, we also discuss some promising hybridizations between TSK fuzzy systems and each of the four algorithms, which could be interesting new research directions. For example:
\begin{enumerate}
\item By making use of the functional equivalence between TSK fuzzy systems and some neural networks, we can design more efficient training algorithms for TSK fuzzy systems.
\item By making use of the functional equivalence between TSK fuzzy systems and MoE, we may be able to achieve a better trade-off between cooperations and competitions of the rules in a TSK fuzzy system.
\item By making use of the functional equivalence between TSK fuzzy systems and CART, we can better initialize a TSK fuzzy system for high-dimensional problems.
\item Inspired by the connections between TSK fuzzy systems and stacking ensemble regression, we may be able to design better stacking models, and increase the generalization ability of a TSK fuzzy model.
\end{enumerate}

The remainder of this paper is organized as follows: Sections~\ref{sect:NN}-\ref{sect:stacking} describe the functional equivalence of TSK fuzzy systems to neural networks, MoE, CART, and stacking ensemble regression, respectively. Section~\ref{sect:conclusion} draws conclusion.

\section{TSK Fuzzy Systems and Neural Networks} \label{sect:NN}

Neural networks have a longer history\footnote{https://cs.stanford.edu/people/eroberts/courses/soco/projects/neural-networks/History/history1.html} than fuzzy systems, and are now at the center stage of machine learning, because of the booming of deep learning \cite{Hinton2006}.

Researchers started to discover in the early 1990s that a TSK fuzzy system can be represented similarly to a neural network \cite{Halgamuge1994,Jang1993,Berenji1992,Buckley1994,Wang1992b}, so that a neural network learning algorithm, such as back-propagation \cite{Bishop1995}, can be used to train it. These fuzzy systems are called \emph{neuro-fuzzy systems} in the literature \cite{Lin1996a}.

\subsection{ANFIS}

\begin{figure}[htbp]         \centering
\includegraphics[width=\linewidth,clip]{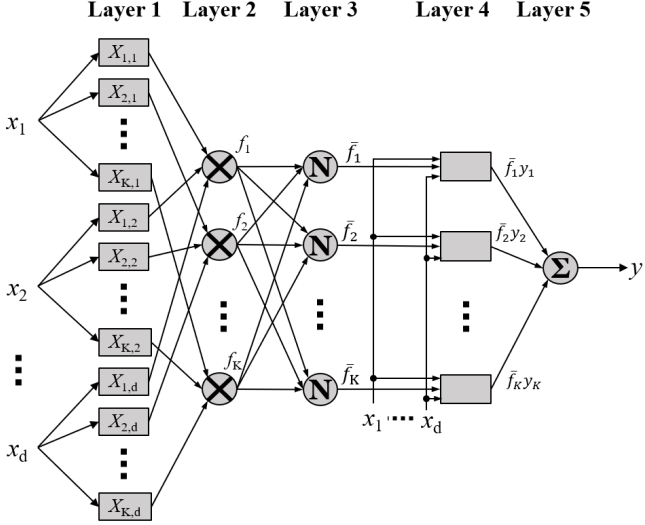}
\caption{The TSK fuzzy system introduced in the Introduction, represented as a 5-layer ANFIS.} \label{fig:ANFIS}
\end{figure}

Among the many variants of neuro-fuzzy systems, the most popular one may be the ANFIS \cite{Jang1993}, which has been cited over 15,000 times on Google Scholar, and implemented in the Matlab Fuzzy Logic Toolbox. The ANFIS structure of the $d$-input one-output TSK fuzzy system, introduced in the Introduction, is shown in Fig.~\ref{fig:ANFIS}. It has five layers:
\begin{itemize}
\item \emph{Layer 1:} The membership grade of $x_i$ on $X_{k,i}$ ($k=1,...,K$; $i=1,...,d$) is computed.
\item \emph{Layer 2:} The firing level of each rule $R_k$ is computed, by multiplying the membership grades of the corresponding rule antecedents.
\item \emph{Layer 3:} The normalized firing levels of the rules are computed, using (\ref{eq:f}).
\item \emph{Layer 4:} Each normalized firing level is multiplied by its corresponding rule consequent.
\item \emph{Layer 5:} The output is computed by (\ref{eq:yTSK}).
\end{itemize}
All parameters of the ANFIS, i.e., the shapes of the MFs and the rule consequents, can be trained by a gradient descent algorithm \cite{Jang1993}. Or, to speed up the training, a more efficient hybrid learning algorithm \cite{Jang1993} can be used. In the forward pass, the antecedent parameters are fixed, the functional signals go forward till Layer~4, and the consequent parameters are optimized by least squares estimation. In the backward pass, the consequent parameters are fixed, the errors propagate backward, and the antecedent parameters are updated by gradient descent.

\subsection{Functional Equivalence between TSK Fuzzy Systems and Radial Basis Functional Networks (RBFN)}

A variant of neural networks, the radial basis function network (RBFN) \cite{Moody1989}, which is a universal approximator \cite{Park1991}, is functionally equivalent to a TSK fuzzy system under certain constraints. A radial basis function is a real-valued function whose value depends only on the distance from a center $\mathbf{c}$, i.e., $f(\mathbf{x},\mathbf{c})=f(\|\mathbf{x}-\mathbf{c}\|)$.

\begin{figure}[htbp]         \centering
\includegraphics[width=.65\linewidth,clip]{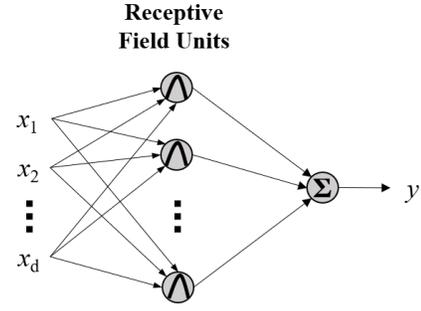}
\caption{The RBFN.} \label{fig:RBF}
\end{figure}

An RBFN \cite{Moody1989} uses local receptive fields, inspired by biological receptive fields, for function mapping. Its diagram is shown in Fig.~\ref{fig:RBF}. For an input $\mathbf{x}=(x_1,...,x_d)$, the output of the $k$th ($k=1,...,K$) receptive field unit, using a Gaussian response function, is:
\begin{align}
f_k(\mathbf{x})=\exp\left(-\frac{\sum_{i=1}^d(x_i-c_{k,i})^2}{\sigma_k^2}\right),
\end{align}
where $c_{k,i}$ is the center of the Gaussian function for $x_i$, and $\sigma_k$ is the common standard deviation of the Gaussian functions.

With the addition of lateral connections (not shown in Fig.~\ref{fig:RBF}) between the receptive field units, the output of the (normalized) RBFN is:
\begin{align}
y(\mathbf{x})=\frac{\sum_{k=1}^K f_k(\mathbf{x})\cdot y_k}{\sum_{k=1}^Kf_k(\mathbf{x})}, \label{eq:yRBFN}
\end{align}
where $y_k$ is a constant output associated with the $k$th receptive field unit\footnote{There is a related machine learning approach called local model networks \cite{Murray-Smith1995}, which can be viewed as a decomposition of a complex nonlinear system into a set of locally accurate sub-models smoothly integrated by their associated basis functions. It replaces the constant output of each receptive unit in an RBFN by a function of the inputs.}.

Jang and Sun \cite{Jang1993a} have shown that a TSK fuzzy system [see (\ref{eq:yTSK})] is functionally equivalent to an RBFN [see (\ref{eq:yRBFN})], if the following constraints are satisfied:
\begin{enumerate}
\item The number of receptive field units equals the number of fuzzy rules.
\item The output of each fuzzy rule is a constant, instead of a function of the inputs.
\item The antecedent MFs of each fuzzy rule are Gaussian functions with the same variance.
\item The product $t$-norm is used to compute the firing level of each rule.
\item The fuzzy system and the RBFN use the same method (i.e., either weighted average or weighted sum) to compute the final output.
\end{enumerate}
Actually, the Gaussian function requirement in Constraint~(3) may not be necessary. When the other four constraints are satisfied, as long as the MFs in the TSK fuzzy system are in the same form as the radial basis functions (not necessarily Gaussian) in the RBFN, the TSK fuzzy system and the RBFN are equivalent.

Hunt \emph{et al.} \cite{Hunt1996} proposed a generalized RBFN, which has the following main features, compared with the above standard RBFN:
\begin{enumerate}
\item A receptive field unit may be connected with only a subset of the inputs, instead of all inputs in the standard RBFN.
\item The output associated with each receptive field unit can be a linear or nonlinear function of the inputs, instead of a constant in the standard RBFN.
\item The Gaussian response functions of the receptive field units can have different variances for different inputs, instead of identical variance in the standard RBFN.
\end{enumerate}
Then, the generalized RBFN is functionally equivalent to a TSK fuzzy system, under the following constraints \cite{Hunt1996}:
\begin{enumerate}
\item The number of receptive field units equals the number of fuzzy rules.
\item The antecedent MFs of each fuzzy rule are Gaussian.
\item The product $t$-norm is used to compute the firing level of each rule.
\item The fuzzy system and the RBFN use the same method (i.e., either weighted average or weighted sum) to compute the final output.
\end{enumerate}
Again, the Gaussian constraint can be relaxed to other radial basis functions.

The training of an RBFN consists of two steps:
\begin{enumerate}
\item Determine the center vectors of the RBFs in the hidden layer. They can be initialized randomly, or through clustering, or by grid partition of the input space.
\item Fit a linear model to the hidden layer's outputs according to some loss function, e.g., the least squares loss in regression.
\end{enumerate}
Because of the functional equivalence between an RBFN and a TSK fuzzy system, the techniques used in Step~(1) above can also be used to initialize the MFs in a TSK fuzzy systems. For example, in evolutionary fuzzy systems design \cite{drwuEAAI2006}, the input MFs are randomly initialized, and a fitness function is used to select the configuration that achieves the best performance. In Yen et al.'s approach \cite{Yen1998} to increase the interpretability of a TSK fuzzy system, the antecedent fuzzy partitions are determined by starting with an oversized number of partitions, and then removing redundant and less important ones using the SVD-QR algorithm \cite{Golub1976}. There have also been many approaches for generating initial fuzzy rule partitions through clustering \cite{Yager1994a,Delgado1997,Chiu1994,Juang1998}. Different clustering algorithms, e.g., mountain clustering \cite{Yager1994a}, fuzzy $c$-means clustering \cite{Delgado1997}, aligned clustering \cite{Juang1998}, etc., have been used.

Moreover, some efficient and effective training approaches for RBFN have been proposed recently, which may also be extended to TSK fuzzy system. For example, multicolumn RBFN \cite{Hoori2018}, which divides a large dataset into smaller subsets using the k-d tree algorithm and then trains an RBFN for each subset, has demonstrated faster speed and higher accuracy than the traditional RBFN. This approach could be extended to TSK fuzzy systems for big data problems.

\subsection{Discussions and Future Research}

As ANFIS is an efficient and popular training algorithm for type-1 TSK fuzzy systems, it is natural to consider whether it can also be used for interval and general type-2 fuzzy systems \cite{Mendel2017}, which have demonstrated better performance than type-1 fuzzy systems in many applications. There have been limited research in this direction \cite{Chen2016,Chen2017a}. Unfortunately, it was found that interval type-2 ANFIS may not outperform type-1 ANFIS. One possible reason is that when the Karnik-Mendel algorithms \cite{Mendel2017} are used in type-reduction of the interval type-2 fuzzy system, the least squares estimator in the interval type-2 ANFIS does not always give the optimal solution, due to the switch point mismatch \cite{Chen2016}. A remedy may be to use an alternative type-reduction and defuzzification approach \cite{drwuCCTFS2013}, which does not involve the switch points, e.g., the Wu-Tan method \cite{drwuTR2005}. This is a direction that we are currently working on.

Many novel approaches have been proposed in the last few years to speed up the training and increase the generalization ability of deep neural networks, e.g., dropout \cite{Srivastava2014}, dropConnect \cite{Wan2013}, and batch normalization \cite{Ioffe2015}. Dropout randomly discards some neurons and their connections during the training. DropConnect randomly sets some connection weights to zero during the training. Batch normalization normalizes the activation of the hidden units, and hence reduces internal covariate shift\footnote{As explained in \cite{Ioffe2015}, internal covariate shift means \emph{``the distribution of each layer's inputs changes during training, as the parameters of the previous layers change. This slows down the training by requiring lower learning rates and careful parameter initialization, and makes it notoriously hard to train models with saturating nonlinearities."}}. Similar concepts may also be used to expedite the training and increase the generalization ability of TSK fuzzy systems. For example, inspired by Dropout, we recently developed a novel dropRule approach \cite{drwuGD2019} for training TSK fuzzy systems, which drops some rules randomly in TSK fuzzy system training.

Although deep learning has achieved great success in numerous applications, its model is essentially a black-box because it is difficult to explain the acquired knowledge or decision rationale. This may hinder it from safety-critical applications such as medical diagnoses. Explainability of deep learning models has attracted a rapidly growing research interest in the past few years. According to Amarasinghe and Manic, there have been two groups of research on this \cite{Amarasinghe2018}: 1) altering the learning algorithms to learn explainable features; and, 2) using additional methods with the standard learning algorithm to explain existing deep learning algorithms. They \cite{Amarasinghe2018} also presented an interesting methodology for linguistically explaining the knowledge a deep neural network classifier has acquired in training, using linguistic summarization \cite{drwuLS2011}, which generates Zadeh fuzzy rules. This work shows a novel and promising application of fuzzy rules in deep learning. Similarly, TSK fuzzy rules could also be used to linguistically explain a deep regression model.

Finally, fuzzy logic and deep learning could be hybridized to take the advantages of both sides \cite{Deng2017,Zheng2017}. \cite{Deng2017} proposed a hierarchical deep neural network that derives information from both fuzzy and neural representations, which are then fused to form the final features for classification. \cite{Zheng2017} proposed a Pythagorean fuzzy deep Boltzmann machine, in which the deep Boltzmann machine parameters are expressed by Pythagorean fuzzy numbers so that each neuron can learn how a feature affects the output both positively and negatively. We expect that more hybridizations like these will emerge in the near future.

\section{TSK Fuzzy Systems and MoE} \label{sect:MoE}

\begin{figure}[htbp]         \centering
\includegraphics[width=.85\linewidth,clip]{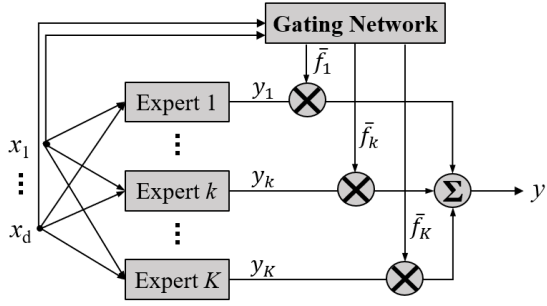}
\caption{Mixture of experts (MoE).} \label{fig:MoE}
\end{figure}

MoE was first proposed by Jacobs \emph{et al.} in 1991 \cite{Jacobs1991}. It is established based on the divide-and-conquer principle, in which the problem space is divided among multiple local experts, supervised by a gating network, as shown in Fig.~\ref{fig:MoE}. MoE trains multiple local experts, each taking care of only a small local region of the problem space; for a new input, the gating network determines which experts should be used for it, and then aggregates the outputs of these experts by a weighted average. MoE models are universal approximators \cite{Nguyen2016}.

\subsection{MoE}

Assume there are $N$ training examples $(\mathbf{x}_n,y_n)$, $n=1,...,N$. The $K$ experts are trained from minimizing the following error\footnote{In practice, transforms of (\ref{eq:errorME}), e.g., $E=-\log \sum_{n=1}^N\sum_{k=1}^K \bar{f}_k(\mathbf{x}_n)\exp{[-\frac{1}{2}(y_n-y_k(\mathbf{x}_n))^2]}$, may be used to speed up the optimization \cite{Jacobs1991}.}:
\begin{align}
E=\sum_{n=1}^N \sum_{k=1}^K \bar{f}_k(\mathbf{x}_n)(y_n-y_k(\mathbf{x}_n))^2, \label{eq:errorME}
\end{align}
where $y_k(\mathbf{x}_n)$ is the output of the $k$th expert for input $\mathbf{x}_n$, and $\bar{f}_k(\mathbf{x}_n)$ is the corresponding normalized weight for the $k$th expert, assigned by the gating network:
\begin{align}
\bar{f}_k(\mathbf{x}_n)=\frac{\exp(f_k(\mathbf{x}_n))}{\sum_{i=1}^K \exp(f_i(\mathbf{x}_n))}, \label{eq:fkMoE}
\end{align}
in which $f_k(\mathbf{x}_n)$ is a tunable function.

Once the training is done, the final output of the MoE is:
\begin{align}
y_{_{MoE}}(\mathbf{x})=\sum_{k=1}^K \bar{f}_k(\mathbf{x}) y_k(\mathbf{x}). \label{eq:yME}
\end{align}


\subsection{Functional Equivalence between TSK Fuzzy Systems and MoE}

It is easy to see that the TSK fuzzy system in (\ref{eq:yTSK}) and the MoE in (\ref{eq:yME}) are conceptually equivalent. More specifically, when the following conditions are satisfied, they are functionally equivalent, i.e., (\ref{eq:yTSK}) is identical to (\ref{eq:yME}):
\begin{enumerate}
\item The TSK fuzzy system uses Gaussian MFs and the product $t$-norm. Assume the Gaussian MF $X_{k,i}$ of the TSK fuzzy system [see (\ref{eq:Rule})] has center $c_{k,i}$ and standard deviation $\sigma_{k,i}$. Then, $f_k(\mathbf{x}_n)$ in (\ref{eq:fkMoE}) of the MoE should be
    \begin{align}
    f_k(\mathbf{x}_n)=-\sum_{i=1}^d \frac{(x_{n,i}-c_{k,i})^2}{2\sigma_{k,i}^2}.
    \end{align}
\item $y_k(\mathbf{x})$ in the TSK fuzzy system is identical to $y_k(\mathbf{x})$ in the MoE, for every $k$.
\end{enumerate}

A few publications \cite{Bersini1997,Andersen1998} have made the connection between TSK fuzzy systems and MoE. The regression function in each rule consequent of the TSK fuzzy system can be viewed as an expert, and the rule antecedents work as the gating network: for each input, they determine how much weight should be assigned to each rule consequent (expert) in the final aggregation. Of course, the experts and gating network in MoE can be constructed by more complex models, such as neural networks and support vector machines \cite{Yuksel2012}, but the structure resemblance remains unchanged.

\subsection{Discussions and Future Research}

Lots of progresses on MoE have been made since it was first proposed in 1991 \cite{Yuksel2012,Masoudnia2014}, e.g., different training algorithms, different gating networks, and different expert models. Since MoE is essentially identical to a TSK fuzzy system, these ideas could also be applied to fuzzy systems, particularly the training algorithms and expert models (the gating network is a little more challenging because in a TSK fuzzy system we always use MFs to perform gating; there is not too much freedom).

First, in training a TSK fuzzy system for regression, the error function is usually defined as:
\begin{align}
E&=\sum_{n=1}^N (y_n-y_{_{TSK}}(\mathbf{x}_n))^2\nonumber \\
&=\sum_{n=1}^N \left[y_n-\sum_{k=1}^K \bar{f}_k(\mathbf{x}_n)y_k(\mathbf{x}_n)\right]^2  \label{eq:ETSK}
\end{align}
However, as pointed out in \cite{Jacobs1991}, \emph{``this error measure compares the desired output with a blend of the outputs of the local experts, so, to minimize the error, each local expert must make its output cancel the residual error that is left by the combined effects of all the other experts. When the weight in one expert change, the residual error changes, and so the error derivatives for all other local experts change."} This strong coupling between the experts facilitates their cooperation, but may lead to a solution in which many experts are used for each input. That's why in training the local experts, the error function is defined as (\ref{eq:errorME}) to facilitate the competition among them. (\ref{eq:errorME}) requires each expert to approximate $y_n$, instead of a residual. Hence, each local expert is not directly affected by other experts (it is indirectly affected by other experts through the gating network, though).

It is thus interesting to study if changing the error function from (\ref{eq:ETSK}) to (\ref{eq:errorME}) in training a TSK fuzzy system can improve its performance, in terms of speed and accuracy. Or, the error function could be a hybridization of (\ref{eq:ETSK}) and (\ref{eq:errorME}), to facilitate both the cooperations and competitions among the local experts, i.e.,
\begin{align}
E=&\sum_{n=1}^N \left[y_n-\sum_{k=1}^K\bar{f}_k(\mathbf{x}_n)y_k(\mathbf{x}_n)\right]^2\nonumber \\
&+\lambda \sum_{n=1}^N\sum_{k=1}^K \bar{f}_k(\mathbf{x}_n) \left[y_n-y_k(\mathbf{x}_n)\right]^2,
\end{align}
where $\lambda$ is a hyper-parameter defining the trade-off between cooperation and competition. This idea was first explored in \cite{Yen1998}, which proposed a TSK fuzzy system design strategy to increase its interpretability, by forcing each rule consequent to be a reasonable local model (the regression function in each rule consequent needs to fit the training data that are covered by the rule antecedent MFs well), and also the overall TSK fuzzy system to be a good global model. However, the algorithms in \cite{Yen1998} are very memory-hungry\footnote{Let $N$ be the number of training examples, $r$ the dimensionality of the input, and $L$ the number of rules. The algorithms in \cite{Yen1998} need to construct matrices in $\mathbb{R}^{NL\times L(r+1)}$ and $\mathbb{R}^{NL\times NL}$, which are hardly scalable.}, and hence may not be applicable when $N$ is large. A more efficient solution to this problem is needed.

Second, when the performance of an initially designed TSK fuzzy system is not satisfactory, there could be two strategies to improve it: 1) increase the number of rules, so that each rule covers a smaller region in the input domain, and hence may better approximate the training examples in that region; and, 2) increase the fitting power (nonlinearity) of the consequent function, so that it can better fit the training examples in its local region. The first strategy is frequently used in practice; however, it can increase the number of parameters of the TSK fuzzy system very rapidly. Juang and Lin \cite{Juang1998} proposed an interesting approach to incrementally add linear terms to the rule consequent to increase its fitting power. However, they only considered linear terms. Inspired by MoE, whose expert models could use complex models like the neural networks and support vector machines \cite{Yuksel2012}, the TSK rule consequents (local experts) could also use more sophisticated models, particularly, support vector regression \cite{Smola2004}, which outperforms simple linear regression in many applications. The feasibility of this idea has been verified in \cite{Juang2010,Komijani2012}.

Third, TSK fuzzy rules could also be used as experts in MoE. For example, Leski \cite{Leski2003} proposed such an approach for classification: each expert model in the MoE was constructed as a TSK fuzzy rule (whose input region was determined by fuzzy $c$-means clustering), and then a gating network was used to aggregate them. This may increase the interpretability of MoE. This idea can also be extended to regression problems.

\section{TSK Fuzzy Systems and CART} \label{sect:CART}

CART \cite{Breiman2017} is a popular and powerful strategy for constructing classification and regression trees. It is a universal approximator \cite{Berry2006}, and has also been used in ensemble learning such as random forests \cite{Breiman2001} and gradient boosting machines \cite{Friedman2001}. This section focuses on regression only.

\subsection{CART}

\begin{figure}[htbp]\centering
\subfigure[]{\label{fig:CART}   \includegraphics[width=.48\linewidth,clip]{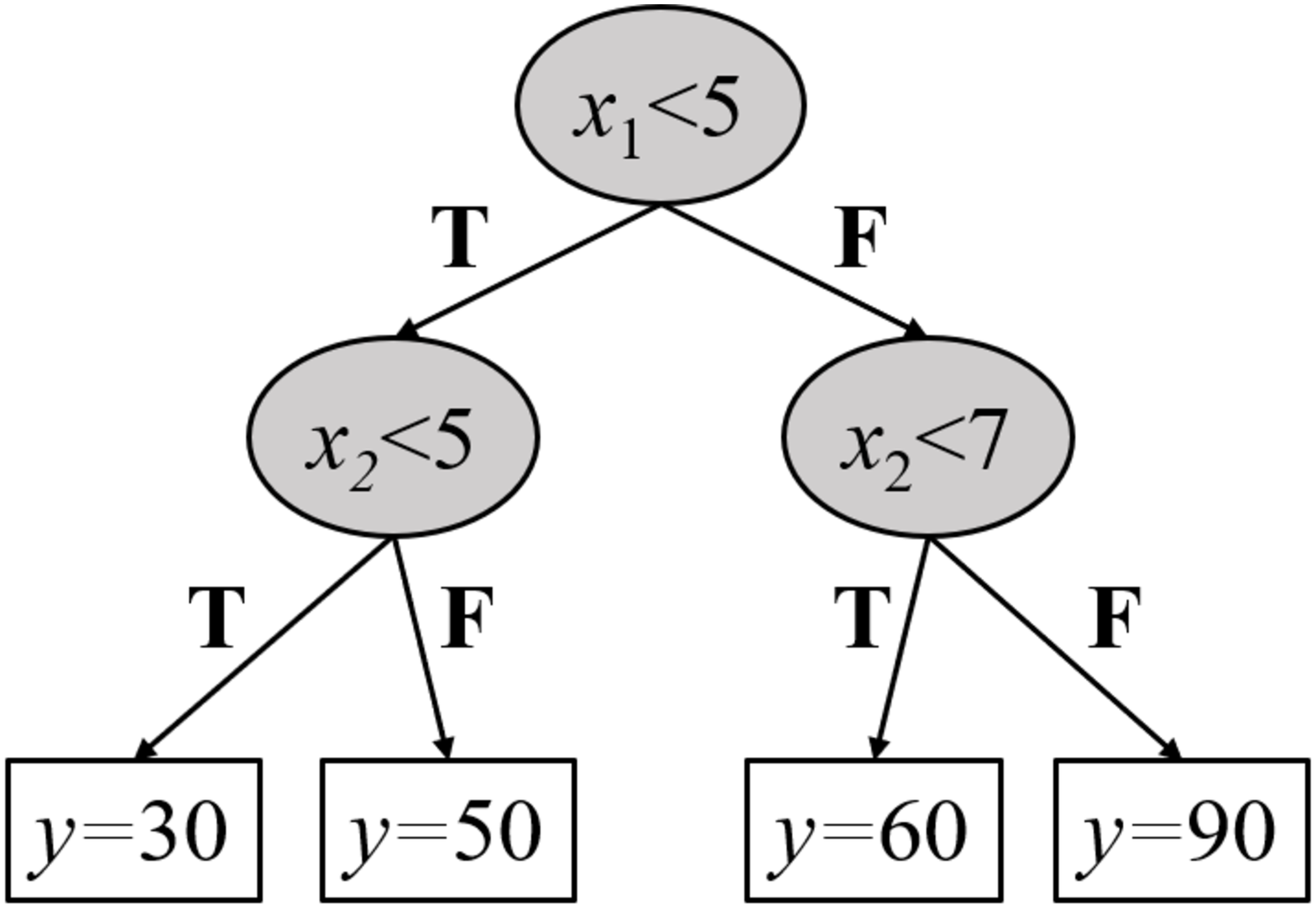}}
\subfigure[]{\label{fig:DT}   \includegraphics[width=.48\linewidth,clip]{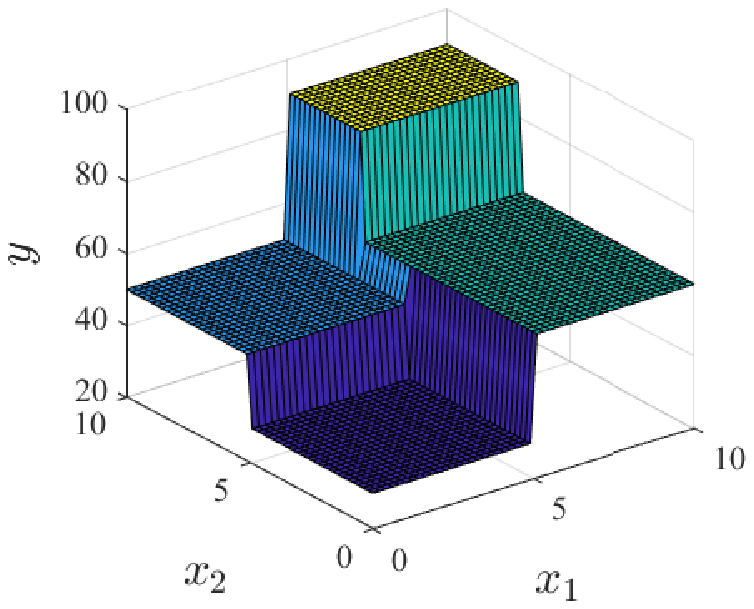}}
\caption{(a) An example of CART for regression; and, (b) its input-output mapping.}
\end{figure}

Assume there are two numerical inputs, $x_1$ and $x_2$, and one output, $y$. An example of CART is shown in Fig.~\ref{fig:CART}. It is constructed by a divide-and-conquer strategy, in which the input space is
partitioned by a hierarchy of Boolean tests into multiple non-overlapping partitions. Each Boolean test corresponds to an internal node of the decision tree. The leaf node (terminal node) is computed as the mean $y$ of all training examples falling into the corresponding partition; thus, CART implements a piecewise constant regression function, as shown in Fig.~\ref{fig:DT}. The route leading to each leaf node can be written as a crisp rule, e.g., If $x_1<5$ and $x_2<5$, then $y=30$. Note that each leaf node can also be a function of the inputs \cite{Loh2014,Alexander1996,Chaudhuri1994,Quinlan1992a,Dobra2002}, instead of a constant. In this way, the implemented regression function is smoother; however, the trees are more difficult to train.

\subsection{Functional Equivalence between TSK Fuzzy Systems and CART}

Both CART and fuzzy systems use rules. The rules in CART are crisp: each input belongs to only one rule, and the output is the leaf node of that rule. On the contrary, the rules in a fuzzy system are fuzzy: each input may fire more than one rules, and the output is a weighted average of these rule consequents.

\begin{figure}[htbp]\centering
\subfigure[]{\label{fig:fuzzyCART}   \includegraphics[width=.48\linewidth,clip]{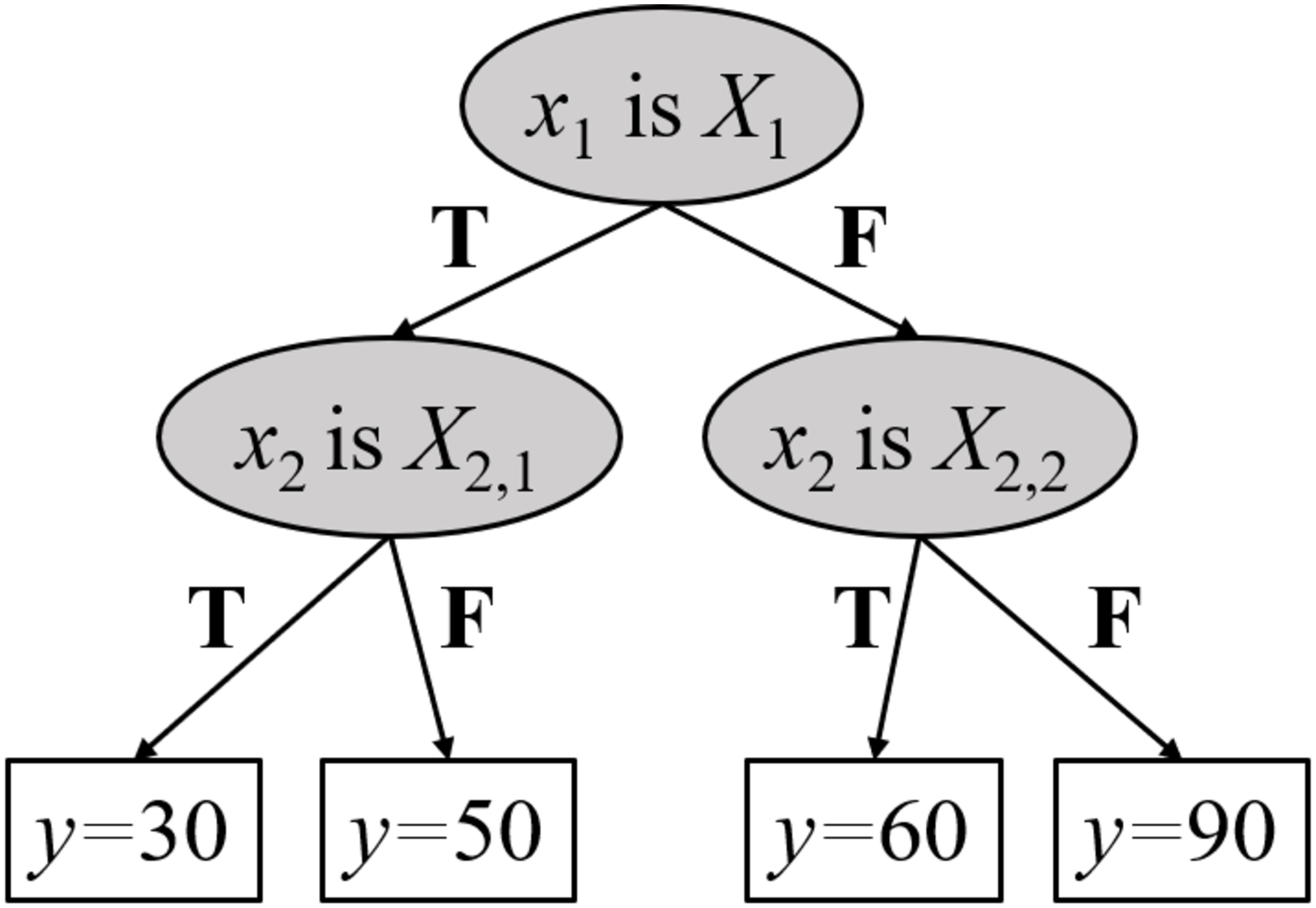}}
\subfigure[]{\label{fig:fDT}   \includegraphics[width=.48\linewidth,clip]{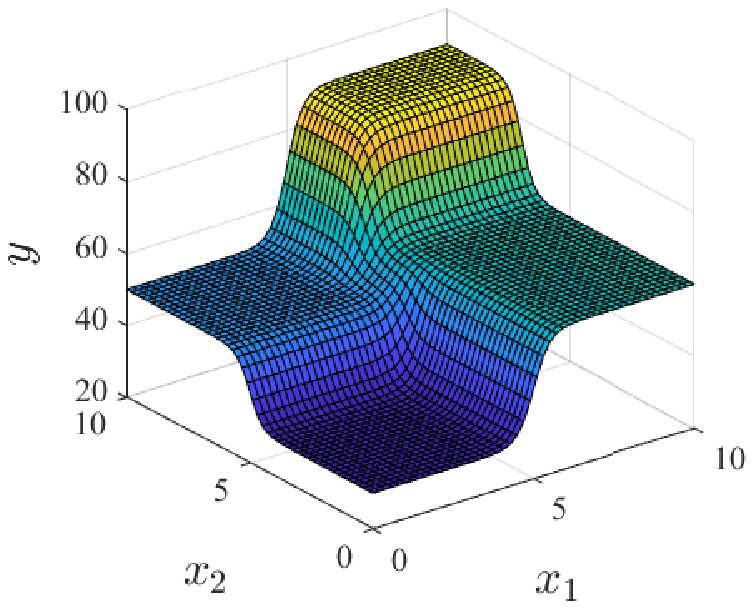}}
\caption{(a) An example of fuzzy CART for regression; and, (b) its input-output mapping.}
\end{figure}

The regression output of a traditional CART has discontinuities, which may be undesirable in practice. So, fuzzy CART, which allows an input to belong to different leaf nodes with different degrees, has been proposed to accommodate this \cite{Chang1977,Jang1994,Janikow1998,Suarez1999}. As pointed out by Suarez and Lutsko \cite{Suarez1999}, \emph{``in regression problems, it is seen that the continuity constraint imposed by the function representation of the fuzzy tree leads to substantial improvements in the quality of the regression and limits the tendency to overfitting."} An example of fuzzy CART for regression is shown in Fig.~\ref{fig:fuzzyCART}, where $X_1$ is a fuzzy set for $x_1$, and $X_{2,1}$ and $X_{2,2}$ are fuzzy sets for $x_2$. Its input-output mapping is shown in Fig.~\ref{fig:fDT}, which is continuous.

Let the $k$th constant leaf node in a fuzzy CART be $y_k$. Then, given an input $\mathbf{x}$, the output of a fuzzy CART is a weighed average of the predictions at all leaves:
\begin{align}
y_{_{fCART}}(\mathbf{x})=\frac{\sum_{k=1}^Kf_k(\mathbf{x})y_k}{\sum_{k=1}^K f_k(\mathbf{x})}, \label{eq:yFuzzyCART}
\end{align}
where $f_k(\mathbf{x})$ is the product of all membership grades in the path to $y_k$. In this way, $y_{_{fCART}}(\mathbf{x})$ is a smooth function. Clearly, $y_{_{fCART}}(\mathbf{x})$ is functionally equivalent to the output of a TSK fuzzy system in (\ref{eq:yTSK1}), when each rule consequent of the fuzzy system is a constant (instead of a function).

Chaudhuri \emph{et al.} \cite{Chaudhuri1994} proposed smoothed and unsmoothed piecewise-polynomial regression trees (SUPPORT), in which each leaf node is a polynomial function of the inputs. The SUPPORT tree is generally much shorter than a traditional CART tree, and hence enjoys better interpretability. The following three-step procedure is used to ensure that its output is smooth:
\begin{enumerate}
\item The input space is recursively partitioned until the data in each partition are adequately fitted by a fixed-order polynomial. Partitioning is guided by analyzing the distributions of the residuals and the cross-validation estimates of the mean squared prediction error.
\item The data within a neighborhood of the $k$th ($k=1,...,K$) partition are fitted by a polynomial $y_k(\mathbf{x})$.
\item The prediction for an input $\mathbf{x}$ is a weighted average of $y_k(\mathbf{x})$, where the weighting function $f_k(\mathbf{x})$ diminishes rapidly to zero outside the $k$th partition.
\end{enumerate}

If the weighting functions are Gaussian-like, i.e.,
\begin{align}
f_k(\mathbf{x})=\exp\left(-\frac{(x_1-m_{k,1})^2}{\sigma_{k,1}^2}-\frac{(x_2-m_{k,2})^2}{\sigma_{k,2}^2}\right),
\end{align}
where $m_{k,i}$ is the mean of the Gaussian function for the $i$th input, and $\sigma_{k,i}$ is the standard deviation, then the output of SUPPORT is:
\begin{align}
y_{_{SUPPORT}}(\mathbf{x})=\frac{\sum_{k=1}^K f_k(\mathbf{x})y_k(\mathbf{x})}{\sum_{k=1}^Kf_k(\mathbf{x})}.
\end{align}
Clearly, $y_{_{SUPPORT}}(\mathbf{x})$ is functionally equivalent to the output of the TSK fuzzy system in (\ref{eq:yTSK1}).

\subsection{Discussions and Future Research}

It is well-known that fuzzy systems are particularly subject to the curse of dimensionality: as the number of features increases, the number of rules may increase exponentially, and the interpretability also decreases quickly as the number of antecedents increases.

CART may offer a solution to this problem. For example, Jang \cite{Jang1994} first performed CART on a regression dataset to roughly estimate the structure of a TSK fuzzy system, i.e., number of MFs in each input domain, and the number of rules. Then, each crisp rule antecedent was converted into a fuzzy set, and consequent to a linear function of the inputs. For example, a crisp rule:
\begin{align*}
R_k: \mbox{ If } x_1 > x_{k,1} \mbox{ and } x_2 <x_{k,2}, \mbox{ then } y_k=c_k
\end{align*}
can be converted to a TSK fuzzy rule:
\begin{align*}
R_k: &\mbox{ If } x_1 \mbox{ is } X_{k,1} \mbox{ and } x_2 \mbox{ is } X_{k,2},\\
& \mbox{ then } y_k=a_kx_1+b_kx_2+c_k',
\end{align*}
where $a_k$, $b_k$ and $c_k'$ are regression coefficients, and $X_{k,1}$ and $X_{k,2}$ are fuzzy sets defined as:
\begin{align}
\mu_{X_{k,1}}(x_1)&=\frac{1}{1+\exp[-\alpha_{k,1}(x_1-x_{k,1})]}\\
\mu_{X_{k,2}}(x_2)&=\frac{1}{1+\exp[\alpha_{k,2}(x_2-x_{k,2})]}
\end{align}
in which $\alpha_{k,1}$ and $\alpha_{k,2}$ are tunable parameters. Once all crisp rules have been converted to TSK fuzzy rules, ANFIS \cite{Jang1993} can be used to optimize the parameters of all rules together, e.g., $a_k$, $b_k$, $c_k'$, $\alpha_{k,1}$, and $\alpha_{k,2}$.

The above TSK fuzzy system design strategy offers at least three advantages:
\begin{enumerate}
\item \emph{Simplicity}. We can prune the CART tree on a high-dimensional dataset to obtain a regression tree with a desired number of leaf nodes, and hence a TSK fuzzy system with a desired number of rules. So, we can directly control the simplicity of the resulted TSK fuzzy system.
\item \emph{Interpretability}. Rules in a traditionally designed fuzzy system usually have the same number of antecedents (which equals the number of inputs), which are difficult to interpret when there are many antecedents. Rules initialized from CART may have different number of antecedents (which are usually smaller than the number of inputs), depending on the depths of the corresponding leaf nodes, i.e., we can extract shorter and more interpretable rules that may not be extractable using a traditional fuzzy system design approach.
\item \emph{Performance}. In a traditional fuzzy system, each input (feature) is independently considered in rule antecedents. However, some variants of CART \cite{Loh2014} allow to split on the linear combinations of the inputs, which is equivalent to using new (usually more informative) features in splitting. These new features are also used by fuzzy rules when they are converted from CART leaf nodes, which may achieve better performance than traditional fuzzy systems.
\end{enumerate}
In summary, initializing TSK fuzzy systems from CART regression trees is a promising solution to high-dimensional problems, and may achieve better interpretability-performance trade-off. Hence, it deserves further research.

\section{Fuzzy System and Stacking} \label{sect:stacking}

Ensemble regression \cite{Zhou2012} improves the regression performance by integrating multiple base models. Stacking may be the simplest supervised ensemble regression approach. Its final regression model is a weighted average of the base models, where the weights are trained from the labeled training data. As long as the base models are universal approximators, the stacking model should also be a universal approximator.

%

\subsection{Stacking}

\begin{figure}[htbp]\centering
\includegraphics[width=\linewidth,clip]{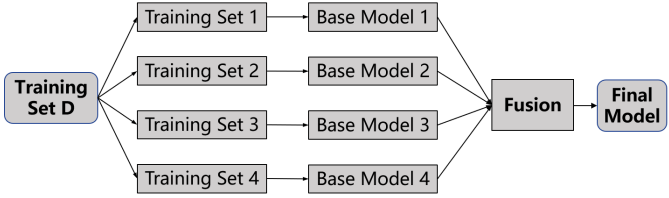}
\caption{Stacking ensemble regression.} \label{fig:Stacking}
\end{figure}

The base models in stacking may be trained from other related tasks or datasets \cite{drwuASECG2019}. However, when there are enough training examples for the problem under consideration, the base models may also be trained directly from them. Fig.~\ref{fig:Stacking} illustrates such a strategy. For a given training dataset, we can re-sample (e.g., using bootstrap \cite{Efron1993}) it multiple times to obtain multiple new training datasets, each of which is slightly different from the original training dataset. Then, a base model can be trained using each re-sampled dataset. These base models could use the same regression algorithm, but different regression algorithms, e.g., LASSO \cite{Tibshirani1996}, ridge regression \cite{Hoerl1970}, support vector regression \cite{Smola2004}, etc., can also be used to increase their diversity. Because the training datasets are different, the trained base models will be different even if the same regression algorithm is used.

Once the $K$ base models are obtained, stacking trains another (linear or nonlinear) regression model to fuse them. Assume the outputs of the $K$ base regression models are $\{y_k\}_{k=1}^K$. Stacking finds a regression model $y=f(y_1,...,y_K)$ on the training dataset to aggregate them.

\subsection{Connections between TSK Fuzzy Systems and Stacking}

A TSK fuzzy system for regression can be viewed as a stacking model. Each rule consequent is a base regression model, and the rule antecedent MFs determine the weights of the base models in stacking. Note that in stacking usually the aggregated output $y$ is a function of $\{y_k\}_{k=1}^K$ only, but in a TSK fuzzy system the aggregation function also depends on the input $\mathbf{x}$, as the weights are computed from them, and change with them. So, a TSK fuzzy system is actually an adaptive stacking regression model.

A key issue in stacking is the partition of the original training dataset, after which a base regression model can be built from each partition. Two commonly used data partition approaches are: 1) bootstrap sampling, where each base model is trained from about 63.2\% unique samples of the original training dataset; and, 2) $k$-fold partition, where each of the $k$ base models is trained from $\frac{100(k-1)}{k}\%$ samples of the original training dataset. Similar dataset partition concepts have also been used to construct or initialize TSK fuzzy systems; however, they may be better interpreted from a possibility point of view, where the possibility (weight) of a sample in constructing a TSK rule is usually a function of the corresponding firing level.

For example, Nozaki \emph{et al.} \cite{Nozaki1997} proposed a simple yet powerful heuristic approach for generating TSK fuzzy rules (whose rule consequents are constants, instead of functions of the inputs) from numerical data. Assume there are $N$ training examples $(x_{n,1},...,x_{n,d},y_n)$, $n=1,...,N$, i.e., the fuzzy system has $d$ inputs and one output. Then, Nozaki \emph{et al.}'s approach consists of the following steps \cite{Nozaki1997}:
\begin{enumerate}
\item Determine how many MFs should be used for each input, and define the shapes of the MFs. Once this is done, the input space is partitioned into several fuzzy regions.
\item Generate a fuzzy rule in the form of:
\begin{align*}
R_k: &\mbox{ If } x_1 \mbox{ is } X_{k,1} \mbox{ and } \cdots \mbox{ and } x_d \mbox{ is } X_{k,d},\\
&  \mbox{ then } y_k = c_k
\end{align*}
in the $k$th fuzzy region, where the MFs $X_{k,i}$ ($i=1,...,d$) have been determined in Step~(1), and
\begin{align}
c_k=\frac{\sum_{n=1}^N\left[\prod_{i=1}^d \mu_{X_{k,i}}(x_{n,i})\right]^\alpha y_n}
{\sum_{n=1}^N\left[\prod_{i=1}^d\mu_{X_{k,i}}(x_{n,i})\right]^\alpha},
\end{align}
in which $\alpha$ is a positive constant. $c_k$ could also be computed using a least squares approach \cite{Nozaki1997}.
\end{enumerate}
Essentially, the above rule-construction approach re-weights each training example in a fuzzy partition using an exponential function of the firing level of the corresponding rule, and then computes a simple base model $y_k=c_k$ from them. The final TSK fuzzy system is an aggregation of all such rules. This is exactly the idea of stacking in Fig.~\ref{fig:Stacking}.

Another example is the local learning part in \cite{Yen1998}, where an approach for constructing a local TSK rule for each rule partition is proposed. Using again the $d$-input one-output example in the Introduction, a local TSK rule is in the form of:
\begin{align*}
R_k: &\mbox{ If } x_1 \mbox{ is } X_{k,1} \mbox{ and } \cdots \mbox{ and } x_d \mbox{ is } X_{k,d},\\
 &\mbox{ then } y_k = \sum_{i=1}^d a_ix_i+b_k.
\end{align*}
Given $X_{k,i}$ ($i=1,...,d$), the weight for each training example $(x_{n,1},...,x_{n,d},y_n)$ is $f_k(\mathbf{x}_n)=\prod_{i=1}^d \mu_{X_{k,i}}(x_{n,i})$, i.e., its firing level of the rule, and then the regression coefficients $a_{k,i}$ and $b_k$ are found from minimizing the following weighted loss:
\begin{align}
E=\sum_{n=1}^Nf_k(\mathbf{x}_n)\left[y_n-\left(\sum_{i=1}^d a_{k,i}x_{n,i}+b_k\right)\right]^2
\end{align}
Each local rule is equivalent to a base model in stacking.

\subsection{Discussions and Future Research}

Traditional stacking assigns each base model a constant weight. As pointed out in the previous subsection, a TSK fuzzy system can be viewed as an adaptive stacking model, because the weights for the base models (rule consequents) change with the inputs. Inspired by this phenomenon, we can design more powerful stacking strategies, by replacing each constant weight by a linear or nonlinear function\footnote{This idea was first used in \cite{Ebrahimpour2011} for \emph{classification}, under the name ``modified stacked generalization." It outperformed traditional stacking.} of the input $\mathbf{x}$. The rationale is that the weight for a base model should be dependent on its performance, whereas its performance is usually related to the location of the input: each base model may perform well in some input regions, but not well in the rest. A well-trained function of $\mathbf{x}$ may be able to reflect the expertise of the corresponding base model, and hence help achieve better aggregation performance.

Moreover, if the weighting functions and the base models are trained simultaneously, then the weighting functions may encourage the base models to cooperate: each focuses on a partition of the input domain, instead of the entire domain in traditional stacking. Even better performance could be expected, than training the base models first and then separately the weighting functions to aggregate them.

Some proven strategies in stacking may also be used to improve the performance of a TSK fuzzy system. For example, regularization is frequently used to increase the generalization of the stacking model \cite{drwuASECG2019, Liu1999}. When LASSO \cite{Tibshirani1996} is used to build the stacking model, $L_1$ regularization is added, and hence some regression coefficients may be zero, i.e., it increases the sparsity of the solution. When ridge regression \cite{Hoerl1970} or support vector regression \cite{Smola2004} is used to build the stacking model, $L_2$ regularization is added, and hence the regression coefficients usually have small magnitudes, i.e., they reduces overfitting. Some new regularization terms, e.g., negative correlation \cite{Liu1999}, can be used to create negatively correlated base models to encourage specialization and cooperation among them. These concepts may also be used in training the rule consequents (base models) of a TSK fuzzy system, and also the antecedent MFs (so that the MFs for the same input are neither too crowded, nor too far away from each other).

\section{Conclusion} \label{sect:conclusion}

TSK fuzzy systems have achieved great success in numerous applications. However, there are still many challenges in designing an optimal TSK fuzzy system, e.g., how to efficiently optimize its parameters, how to balance the trade-off between cooperations and competitions among the rules, how to overcome the curse of dimensionality, how to improve its performance without adding too many parameters, etc. Literature has shown that by making appropriate connections between fuzzy systems and other machine learning approaches, good practices from other domains may be used to improve the fuzzy systems, and vice versa.

This paper has given an overview on the functional equivalence between TSK fuzzy systems and four classic machine learning approaches -- neural networks, MoE, CART, and stacking -- for regression problems. We also pointed out some promising new research directions, inspired by the functional equivalence, that could lead to solutions to the aforementioned problems. For example, by making use of the functional equivalence between TSK fuzzy systems and some neural networks, we can design more efficient training algorithms for TSK fuzzy systems; by making use of the functional equivalence between TSK fuzzy systems and MoE, we may be able to achieve a better trade-off between cooperations and competitions of the rules in a TSK fuzzy system; by making use of the functional equivalence between TSK fuzzy systems and CART, we can better initialize a TSK fuzzy system to deal with the curse of dimensionality; and, inspired by the connections between TSK fuzzy systems and stacking, we may design better stacking models, and increase the generalization of a TSK fuzzy model.

To our knowledge, this paper is so far the most comprehensive overview on the connections between fuzzy systems and other popular machine learning approaches, and hopefully will stimulate more hybridization between different machine learning algorithms.


\end{document}